\documentclass{article}

\usepackage{arxiv}

\usepackage[utf8]{inputenc}
\usepackage[T1]{fontenc}
\usepackage{hyperref}
\usepackage{url}
\usepackage{booktabs}
\usepackage{amsfonts}
\usepackage{amsmath}
\usepackage{bm}
\usepackage{nicefrac}
\usepackage{microtype}
\usepackage{graphicx}
\usepackage{natbib}
\usepackage{doi}
\usepackage{multirow}
\usepackage{siunitx}
\usepackage[nolist,nohyperlinks]{acronym}

\usepackage{authblk}

\renewcommand{\headeright}{}
\renewcommand{\undertitle}{}

\renewcommand{\shorttitle}{Embedding $\ell_1$-regression for temporal structure}

\title{Embedding interpretable $\ell_1$-regression into neural networks for uncovering temporal structure in cell imaging}

\author[1]{Fabian Kabus}
\author[1]{Maren Hackenberg}
\author[2]{Julia Hindel}
\author[3]{Thibault Cholvin}
\author[3]{Antje Kilias}
\author[2]{Thomas Brox}
\author[2]{Abhinav Valada}
\author[3]{Marlene Bartos}
\author[1,4,5]{Harald Binder}

\affil[1]{Institute of Medical Biometry and Statistics (IMBI), Faculty of Medicine and Medical Center, University of Freiburg, Germany}
\affil[2]{Department of Computer Science, University of Freiburg, Germany}
\affil[3]{Institute of Physiology I, Faculty of Medicine, University of Freiburg, Germany}
\affil[4]{Freiburg Center for Data Analysis, Modeling and AI (FDMAI), University of Freiburg, Germany}
\affil[5]{Centre for Integrative Biological Signalling Studies (CIBSS), University of Freiburg, Germany}

\begin{document}
\maketitle

\begin{acronym}[LARS]
\acrodef{ReLU}{rectified linear unit}
\acrodef{SGD}{stochastic gradient descent}
\acrodef{LARS}{least angle regression}
\acrodef{VAR}{vector autoregressive model}
\acrodef{ROI}{region of interest}
\end{acronym}

\begin{abstract}
While artificial neural networks excel in unsupervised learning of non-sparse structure, classical statistical regression techniques offer better interpretability, in particular when sparseness is enforced by $\ell_1$ regularization, enabling identification of which factors drive observed dynamics.
We investigate how these two types of approaches can be optimally combined, exemplarily considering two-photon calcium imaging data where sparse autoregressive dynamics are to be extracted.
We propose embedding a vector autoregressive (VAR) model as an interpretable regression technique into a convolutional autoencoder, which provides dimension reduction for tractable temporal modeling.
A skip connection separately addresses non-sparse static spatial information, selectively channeling sparse structure into the $\ell_1$-regularized VAR.
$\ell_1$-estimation of regression parameters is enabled by differentiating through the piecewise linear solution path.
This is contrasted with approaches where the autoencoder does not adapt to the VAR model.
Having an embedded statistical model also enables a testing approach for comparing temporal sequences from the same observational unit.
Additionally, contribution maps visualize which spatial regions drive the learned dynamics.

\end{abstract}


\section{Introduction}
Integrating neural networks with interpretable statistical models is a promising approach that combines the strengths of both paradigms \citep{shlezinger2023modelbased}. 
Neural networks excel in capturing complex non-sparse patterns in high-dimensional data, making them well suited for tasks such as visual feature extraction and dimensionality reduction, e.g. using convolutional networks and autoencoders. 
In contrast, classical statistical models typically provide interpretability through their theoretical guarantees. 
For instance, autoregressive time series models provide interpretable parameter estimates besides predictions \citep{lutkepohl2005new}.
The most important factors can be obtained by fitting parameters under $\ell_1$-regularization, as popularized by the Lasso approach \citep{tibshirani1996regression}, which can provide a small set of non-zero parameter estimates.
This approach has established itself for a broad class of regression settings and, in particular, enables the identification of sparse patterns in time series models \citep{cavalcante2017lasso}.

Therefore, a combination of a neural network with $\ell_1$-regression could address non-sparse complex structure with the former and sparse, interpretable structure with the latter.
However, this raises the question of how to best direct each approach at the types of structure in a dataset it can excel at.
In addition, joint training is nontrivial, due to different parameter estimation techniques for both types of approaches.
Neural network training typically relies on \ac{SGD} or its variants, e.g. \citep{kingma2015adam}, taking a heuristic approach for finding a local optimum.
In contrast, parameter estimation for classical statistical models often relies on closed-form solutions or small-scale convex optimization, which typically guarantees to find a global optimum under certain assumptions \citep{casella2002statistical, boyd2004convex}. To leverage the advantages of both types of models, we propose a hybrid estimation approach, illustrated on a combination of a convolutional autoencoder neural network with an $\ell_1$ regularized \ac{VAR} model, trained end-to-end.
As an exemplary application setting, we consider extracting sparse temporal patterns from biomedical neural cell imaging videos. There, we also demonstrate how embedding a statistical model into a neural network architecture enables statistical inference.

Specifically, we apply our convolutional encoder to each frame of video to reduce the dimensionality and map time-varying signals into latent space.
Subsequently, an $\ell_1$-regularized \ac{VAR} provides sparse parameter estimates and describes the latent temporal dynamics.
We propose that this approach is valuable in domains requiring both feature extraction and interpretable temporal modeling.
Similar approaches have already been investigated, e.g., in biomedical applications, where longitudinal data often benefit from dimension reduction combined with interpretable temporal models. 
For example, ordinary differential equations \citep{rubanova2019latent, alaa2022icenode, kober2022individualizing}, Gaussian processes \citep{casale2018gaussian, fortuin2020gpvae, ramchandran2021longitudinal}, mixed-effects models \citep{couronne2021longitudinal, sauty2022progression} have been integrated into variational autoencoder frameworks to capture temporal dynamics in a low-dimensional latent space.
Moreover, general linear state-space models have been combined with deep generative models, including probabilistic approaches like deep Kalman filters \citep{krishnan2017structured} and control-oriented frameworks such as \cite{watter2015embed} that assume locally linear dynamics.
However, these approaches typically do not emphasize sparse structure obtained via $\ell_1$-regularization, nor do they leverage classical statistical estimation techniques with established convergence guarantees.

For instance, a naive approach for obtaining parameter estimates for both, the neural networks and the \ac{VAR} model, would be to train the components sequentially, i.e. first the autoencoder, followed by fitting the \ac{VAR} model on the resulting latent space. 
However, disjoint training is prone to sub-optimal results. Due to the non-convex nature of deep neural networks, the optimization may converge to one of many distinct local minima \citep{choromanska2015loss,li2018visualizing}. While these minima may perform equally well for reconstruction, they generate vastly different latent representations. Without feedback from the temporal model, the autoencoder is likely to settle into a minimum where the latent dynamics are not easily captured by a \ac{VAR} process.
Multi-task learning, as an alternative paradigm, would suggest adding the \ac{VAR} loss and $\ell_1$-penalty to the reconstruction loss of the autoencoder.
However, simply combining loss terms in a weighted sum can be problematic.
First, gradients of different objectives might conflict \citep{sener2018multitask}.
Second, the relative scales of those objectives may require extensive hyperparameter tuning \citep{chen2018gradnorm}.
Moreover, this approach would not make use of well-established estimation approaches for $\ell_1$-regularized regression with known convergence properties, such as coordinate descent \citep{wu2008coordinate} and \ac{LARS} \citep{efron2004least}.

In our proposed hybrid approach, these properties can be maintained by unrolling the fitting procedure of the regression model to allow for the use of automatic differentiation techniques \citep{baydin2018automatic} that can provide end-to-end gradients. 
Similar ideas have been explored in domains such as Alternating Direction Method of Multipliers (ADMM) unrolling \citep{yang2016deep} and differentiable ODE solvers \citep{chen2018neural}. For specific regression models, several solvers exist that can implement $\ell_1$-regularization, including coordinate descent \citep{wu2008coordinate}, Fast Iterative Shrinkage-Thresholding Algorithms (FISTA) \citep{beck2009fast}, ADMM \citep{boyd2011distributed} and \ac{LARS} \citep{efron2004least}. 
We illustrate how to adapt the \ac{LARS} approach, motivated by its stability in the face of correlated features \citep{efron2004least} and its efficiency in tracing the regularization path, which helps to mitigate vanishing gradients in unrolled architectures \citep{bengio1994learning}.
By formulating an end-to-end differentiable procedure, where gradients from the statistical model fitting process flow back into the encoder network of the autoencoder, the encoder is enabled to learn representations specifically tailored to facilitate sparse interpretable temporal \ac{VAR} modeling.
To ensure that each component receives the representation it can process effectively, we propose an architectural design that separates static spatial patterns from dynamic temporal structure, routing the former directly to the reconstruction while channeling the latter through the encoder and \ac{VAR} model.

In Section \ref{sec:methods} we detail our proposed hybrid architecture, including a skip connection design for channeling time-varying patterns towards the \ac{VAR} model, and the proposed end-to-end training approach via a differentiable \ac{LARS} algorithm. In addition, we propose a statistical testing procedure based on the \ac{VAR} model, as well as contribution maps that localize the learned dynamics in the original image space. Inspired by concept activation vectors \citep{kim2018interpretability}, which project learned representations back to human-interpretable concepts, our contribution maps project the sparse \ac{VAR} coefficients back to the input image space, revealing which spatial locations drive the learned temporal dynamics. Section \ref{sec:experiments} evaluates the proposed approach in an application to two-photon imaging data \citep{sylte2025coordinated}, a biomedical video scenario where high spatial complexity and temporal patterns of neuronal firing demand an interpretable statistical model. There, we particularly highlight the gained interpretability provided by the proposed approach.
In Section \ref{sec:discussion}, we discuss how the approach might be transferred to other settings where a combination of $\ell_1$-regularized regression and neural networks might be useful. 

\section{Methods}
\label{sec:methods}

\subsection{Channeling temporal structure into a VAR model}
\label{sec:skip_connection}

\begin{figure}[!htbp]
    \includegraphics[width=1\textwidth]{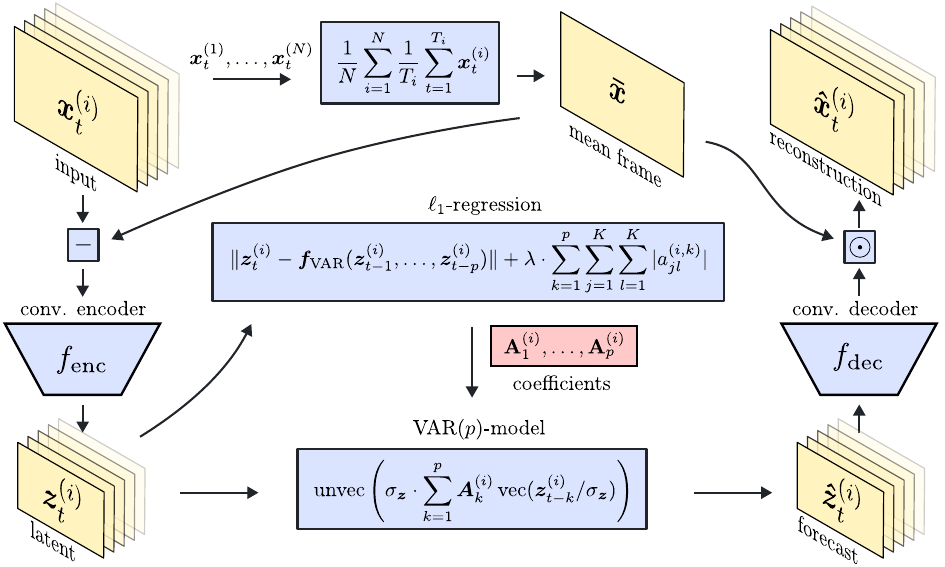}
    \caption{
        High-level overview of the sparse spatiotemporal dimension reduction in the end-to-end model.
        The frames $\bm{x}_t^{(i)}$ for all time series are first aggregated over time to form a mean frame $\bar{\bm{x}}$, which captures the static structure.
        The dynamic component, $\bm{x}_t^{(i)} - \bar{\bm{x}}$, is encoded into a latent representation $\bm{z}_t^{(i)}$.
        The latent representation $\operatorname{vec}(\bm{z}_t^{(i)})$ is modeled using a sparse vector autoregressive (VAR) model of order $p$, which forecasts $\hat{\bm{z}}_t^{(i)}$.
        The coefficient matrices $\bm{A}_1^{(i)}, \ldots, \bm{A}_p^{(i)}$ are fit using $\ell_1$-regression from scratch in each forward pass and they contain the learned spatiotemporal relationships.
        Finally, the decoder reconstructs the frame $\hat{\bm{x}}_t^{(i)}$ from $\hat{\bm{z}}_t^{(i)}$ and the static mean frame $\bar{\bm{x}}$, which is reintroduced via a skip connection.
    }
    \label{fig:model_high_level}
\end{figure}

We consider a dataset of $N$ independent time series, where the $i$-th time series is a sequence of high-dimensional frames $\{\bm{x}_t^{(i)}\}_{t=1}^{T_i}$, with $\bm{x}_t^{(i)} \in \mathbb{R}^{H \times W}$.
The focus in modeling such data is on the dynamic structure, but there might also be significant static content obscuring the temporal information. Therefore, we suggest separating the static information from temporal patterns, as illustrated in Figure \ref{fig:model_high_level}. 
We train model parameters via a reconstruction task that is constrained by a \ac{VAR} model that connects frames of the time series. The original data $\bm{x}_t^{(i)}$ is reconstructed as
$$
\hat{\bm{x}}_t^{(i)} = \left(f_{\text{dec}}\left(f_{\text{VAR}}\left(f_{\text{enc}}(\bm{x}_{t-1}^{(i)} - \bar{\bm{x}}), \ldots, f_{\text{enc}}(\bm{x}_{t-p}^{(i)} - \bar{\bm{x}})\right)\right)+\bm{1}\right) \odot \bar{\bm{x}}  \,\,\,\,\,\,\,\,\, t=p+1,\ldots,T, 
$$
where $\bm{1}$ is a tensor of ones and $\odot$ denotes the Hadamard product, directly connecting the static component $\bar{\bm{x}}$ to the reconstruction, thus implementing a skip connection, where information can bypass the encoder neural network $f_{\text{enc}}: \mathbb{R}^{H \times W} \to  \mathbb{R}^{H' \times W'}$, the \ac{VAR} model $f_{\text{VAR}}: \mathbb{R}^{p \times H' \times W'} \to  \mathbb{R}^{H' \times W'}$, and the decoder $f_{\text{dec}}: \mathbb{R}^{H' \times W'} \to  \mathbb{R}^{H \times W}$. The static component is defined as 
$$\bar{\bm{x}} = \frac{1}{N} \sum_{i=1}^{N} \frac{1}{T_i} \sum_{t=1}^{T_i} \bm{x}_t^{(i)},$$
i.e. the mean over all frames in the dataset. This mean frame captures the non-sparse, time-invariant structure common to all time series and is subtracted from each input frame $\bm{x}_t^{(i)}$ before passing it to the encoder $f_{\text{enc}}$ for obtaining a latent representation
$$\bm{z}_t^{(i)} = f_{\text{enc}}(\bm{x}_{t}^{(i)} - \bar{\bm{x}}) \,\,\,\,\,\,\,\,\, t=1,\ldots,T.$$
This optimizes the latent space exclusively for modeling the time-varying aspects of the data. The latent values of the current frame are predicted based on the previous $p$ frames. We interpret the latent representation as a multivariate time series and apply a \ac{VAR} model of order $p$ with parameters $ \bm{A}_k^{(i)} \in \mathbb{R}^{K \times K}, k=1,\ldots,p$, to capture its dynamics
$$
f_{\text{VAR}}(\bm{z}_{t-1}^{(i)},\ldots,\bm{z}_{t-p}^{(i)}) = \operatorname{unvec}\left(\sigma_{\bm{z}} \cdot \sum_{k=1}^p \bm{A}_{k}^{(i)} \operatorname{vec}(\bm{z}_{t-k}^{(i)}/\sigma_{\bm{z}})\right),
$$
where the function $\operatorname{vec}: \mathbb{R}^{H' \times W'} \to \mathbb{R}^{K}, K=H'\cdot W'$ flattens the latent representation into a vector and the function $\operatorname{unvec}: \mathbb{R}^{K} \to \mathbb{R}^{H' \times W'}$ reshapes the autoregressive output back into a matrix corresponding to the latent representation. To ensure that the \ac{VAR} model operates on data with a consistent scale, the input from the latent representation is normalized by the standard deviation $\sigma_{\bm{z}}$, computed across all latent vectors in the time series $i$.

From the prediction of the VAR model, the decoder $f_{\text{dec}}$ reconstructs the dynamic component of the frame, which is then combined with the static component provided by the skip connection.
The decoder outputs a residual map that modulates the static component to restore the original frame.
By feeding the \ac{VAR} forecast, rather than the original latent state, to the decoder, reconstruction performance depends on the ability of the \ac{VAR} model to accurately predict the latent dynamics.

We use an encoder that comprises three sequential convolutional blocks to downsample the spatial resolution.
Each block consists of a $3 \times 3$ convolutional layer (32 channels), a leaky \ac{ReLU} activation, batch normalization, and a $2 \times 2$ max-pooling layer.
After the final pooling stage, a $1 \times 1$ convolution collapses the feature channels to produce the latent representation $\bm{z}_t^{(i)}$.
The decoder mirrors the encoder's architecture.
It begins with a $1 \times 1$ convolution to expand the channel dimension, followed by three sequential upsampling blocks.
Each decoder block uses max-unpooling with kernel size 2 , leveraging indices from the corresponding encoder max-pooling layer to preserve spatial details. This operation is followed by a $3 \times 3$ convolution, batch normalization, and a leaky \ac{ReLU} normalization.

For obtaining the parameters of the encoder and decoder, we use stochastic gradient descent (SGD) across all $N$ time series.
A single encoder and decoder are shared across all time series, making them comparable at the level of the latent representation.
In contrast, we allow for an individual set of parameters in the VAR model for each time series $i$.
During training, each time series is divided into non-overlapping subsequences of a fixed length, and these subsequences serve as mini-batches for the optimizer.
This approach allows the model to learn from temporal context within each subsequence while permitting efficient batched computation.
The encoder and decoder are trained to minimize the reconstruction loss
$\mathcal{L}_{\text{rec}} = \frac{1}{N} \sum_{i=1}^{N} \frac{1}{T_i H W} \sum_{t=1}^{T_i} \|\bm{x}_t^{(i)} - \hat{\bm{x}}_t^{(i)}\|^2.$

\subsection{Differentiating through least angle regression for the VAR model}
\label{sec:lars}

For estimating the parameters of the VAR model, we consider the $\ell_1$-regularized loss
$$
\|\bm{z}_{t}^{(i)} - f_{\text{VAR}}(\bm{z}_{t-1}^{(i)},\ldots,\bm{z}_{t-p}^{(i)})\| + \lambda\cdot \sum_{k=1}^p \sum_{j=1}^K \sum_{l=1}^K | a_{jl}^{(i,k)} |,
$$
where $a_{jl}^{(i,k)}$ denotes the elements of $A_k^{(i)}$, and $\lambda$ is a tuning parameter that regulates the regularization strength and thus the number of non-zero parameter estimates.

Materializing the complete design matrix for the vectorized \ac{VAR} model would be computationally prohibitive for high-dimensional latent spaces.
To overcome this computational bottleneck, our implementation exploits the sparsity-inducing nature of the \ac{LARS} algorithm.
Instead of constructing the full design matrix, we compute matrix-vector products on-the-fly only for the small subset of active coefficients at each step of the algorithm.
This makes the fitting process tractable.

For estimating the parameters of  $f_{\text{VAR}}$, we aim for end-to-end trainability together with $f_{\text{enc}}$ and $f_{\text{dec}}$, which requires that gradients from the final reconstruction loss propagate back through the entire architecture, including the VAR parameter fitting process.
To achieve this, we fit the \ac{VAR} parameters from scratch using $\ell_1$-regression within each forward pass of the network.
This learning procedure optimizes the encoder to produce a latent representation whose dynamics are well described by a sparse linear model.

Differentiating through $\ell_1$ solvers is, however, nontrivial.
First, the $\ell_1$ penalty is not differentiable at zero, and iterative solvers typically involve branching logic (thresholding, sign checks) that obstructs smooth backpropagation \citep{haselimashhadi2019differentiable, bolte2021nonsmooth,bertrand2020implicit}.
In practice, such non-differentiable operators do not necessarily preclude gradient descent, e.g. the \ac{ReLU} activation function and frequently used $\min$ functions are similar in this respect, and modern AD frameworks can propagate subgradients, but they can still create instability when embedded in larger optimization loops \citep{blondel2025elements}.
A more serious issue arises when iterative algorithms are unrolled into a computational graph: each solver iteration corresponds to another layer in the graph, which can easily run into vanishing or exploding gradient problems when backpropagating through many steps \citep{bengio1994learning,bertrand2020implicit}.

To address these issues, we use the least angle regression (\ac{LARS}) algorithm \citep{efron2004least}, which traces the $\ell_1$ solution by a piecewise-linear homotopy in the regularization tuning parameter~$\lambda$.
Even if it requires many steps, each step is a simple geometry event (one variable entering or leaving the active set) with closed-form coefficient updates.
This yields a more direct computation for differentiation: gradients follow linear segments with kinks only at discrete active-set events.
By contrast, pathwise coordinate‑descent solvers such as implemented in glmnet \citep{friedman2010regularization} perform many sequential updates across variables and along a grid of $\lambda$ values. 

In practice, we observe numerical instability during backpropagation through the \ac{LARS} step that selects the move size $\hat{\gamma}$.
Instability arises when candidate step computations become nearly singular. To prevent this, we add a small sign-aware constant $\epsilon_{\gamma} > 0$ to the relevant terms before selecting the step.
This simple regularization avoids near-division-by-zero and stabilizes gradients without otherwise affecting the path.

\subsection{Statistical testing of group differences in VAR coefficients}
\label{sec:methods_statistical_testing}
To quantitatively assess whether the dynamics learned by the model differ between groups of time series, we employ a statistical test based on swapping the fitted \ac{VAR} coefficients.
For a latent time series $\bm{z}^{(s)}$ with \ac{VAR} coefficients $\{\bm{A}_k^{(s)}\}_{k=1}^p$, we define the prediction error with swapped coefficients $\{\bm{A}_k^{(r)}\}_{k=1}^p$ and the corresponding error differential as
$$
\mathcal{E}_{sr} = \frac{1}{(T_s-p) K} \sum_{t=p+1}^{T_s} \left\|\bm{z}_t^{(s)} - f_{\text{VAR}}^{(r)}(\bm{z}_{t-1}^{(s)},\ldots,\bm{z}_{t-p}^{(s)})\right\|^2, \quad \Delta_{sr} = \mathcal{E}_{sr} - \mathcal{E}_{ss},
$$
where $f_{\text{VAR}}^{(r)}$ denotes the \ac{VAR} function with coefficients $\{\bm{A}_k^{(r)}\}_{k=1}^p$.
The central idea is to test whether swapping coefficients between groups leads to a significantly larger drop in prediction accuracy than swapping them within the same group.
Consider two disjoint groups of time series with index sets $G, H \subseteq \{1, \ldots, N\}$.
We compute these differentials for all intra-group swaps (pairs within $G$ or within $H$) and inter-group swaps (pairs across $G$ and $H$), yielding distributions $\{\Delta_{sr}\}_{\text{intra}}$ and $\{\Delta_{sr}\}_{\text{inter}}$.
Under the null hypothesis of identical group dynamics, these distributions should coincide.
We suggest to test whether $\{\Delta_{sr}\}_{\text{inter}}$ is stochastically greater than $\{\Delta_{sr}\}_{\text{intra}}$ using a one-sided Wilcoxon rank-sum test \citep{wilcoxon1945individual}.

\subsection{Localizing dynamical differences via contribution maps}
\label{sec:methods_visualization}

While the coefficient-swapping test detects global differences, it does not reveal their spatial origin.
As a post-hoc analysis to identify where these differences manifest, we project the learned \ac{VAR} structure back into image space.
We define an influence vector $\bm{c}^{(i)} \in \mathbb{R}^{K}$ and map it to a visual contribution map via
$$
\bm{\Omega}^{(i)} = \left(f_{\text{dec}}(\gamma_{\text{viz}} \cdot \operatorname{unvec}(\bm{c}^{(i)})) + \bm{1}\right) \odot \bar{\bm{x}},
$$
where $\gamma_{\text{viz}}$ is a visualization gain factor.
The influence vector for each time series $i$ captures the total outflow of each latent variable across all lags by aggregating the fitted \ac{VAR} coefficients as $\bm{c}^{(i)}_j = \sum_{k=1}^{p} \sum_{l=1, l \neq j}^{K} \left|a_{jl}^{(i,k)}\right|$, excluding autocorrelation, with absolute values ensuring that both excitatory and inhibitory influences contribute positively to the influence score.

To confirm that this aggregation retains the discriminative information captured by the full \ac{VAR} coefficient matrices, we apply a test analogous to the global coefficient-swapping approach: We compute pairwise squared Euclidean distances $d_{sr} = \|\bm{c}^{(s)} - \bm{c}^{(r)}\|^2$ between influence vectors and test whether inter-group distances exceed intra-group distances using a one-sided Wilcoxon rank-sum test.
A significant result provides an additional indication that the contribution maps reflect meaningful dynamical distinctions rather than noise.

\section{Results}
\label{sec:experiments}
\subsection{Dataset and preparation}
\label{sec:dataset}
To demonstrate our method's applicability, we used it to analyze a real-world dataset. 
The data consists of two-photon calcium imaging recordings of a mouse brain, originally acquired by \citet{sylte2025coordinated}.
The goal is to model neural dynamics, which is an exemplary application for the general problem of modeling sparse, localized dynamics in video data with significant background noise and sensor artifacts.

The recordings exhibit challenges typical of in vivo two-photon imaging, as comprehensively described in reviews and computational frameworks \citep{grienberger2022twophoton,pachitariu2017suite2p,pnevmatikakis2019analysis}.
The signal mostly consists of a structured background originated from tissue autofluorescence, neuropil fluorescence, and systematic illumination or scan-pattern artifacts, which is predominantly static over time.
Superimposed on this background is shot noise from photon statistics, producing unstructured high-frequency fluctuations that can obscure weak events and set fundamental sensitivity trade-offs.
In addition, the planar recording inevitably captures some contribution from out-of-focus structures, leading to a quasi-static haze that further complicates interpretation.
Finally, signals from individual neurons are often contaminated by source overlap and local mixture with surrounding neuropil, causing genuine transients to be blended with background activity.

Distinguishing true, localized transients from this combination of static structure and stochastic noise is therefore non-trivial.
Our architecture reflects this decomposition: the skip path routes static or quasi-static content around the encoder, while end-to-end training of the latent dynamics encourages the model to explain structured, time-localized changes in the latent space and attributes unstructured noise to the reconstruction loss, thereby improving the separability of transients from noise.

The dataset comprises $40$ video runs, captured during an experiment where a mouse navigates one of two virtual environments, denoted F (familiar) and N (novel).
These conditions were used to investigate differences in the temporal dynamics of neurons during spatial navigation.
The runs were acquired in a repeating pattern of five from condition F followed by five from condition N, resulting in 20 F runs and 20 N runs. Each run is a time-series of frames, with lengths ranging from $306$ to $1206$ frames (mean $434.8 \pm 160.9$).

We preprocess the dataset before feeding the frames into the encoder. Each video sequence is divided into non-overlapping subsequences of window size $64$. Further, we center crop the 2D frames to a resolution of $768 \times 512$ and downscaled them with nearest neighbor interpolation to $192 \times 128$. 
Finally, the pixel values were normalized to the range $[0, 1]$.
We use the Adam optimizer with a learning rate $\alpha = 0.001$ for $2$ epochs, the \ac{VAR} model with lag $p = 5$ frames, and $\ell_1$ penalties $\lambda \in \{0.005, 0.01, 0.02\}$.

\subsection{The skip connection improves signal-to-noise ratio in the latent space}
\label{sec:results_skip_connection}
\begin{figure}[!tb]
    \includegraphics[width=1\textwidth]{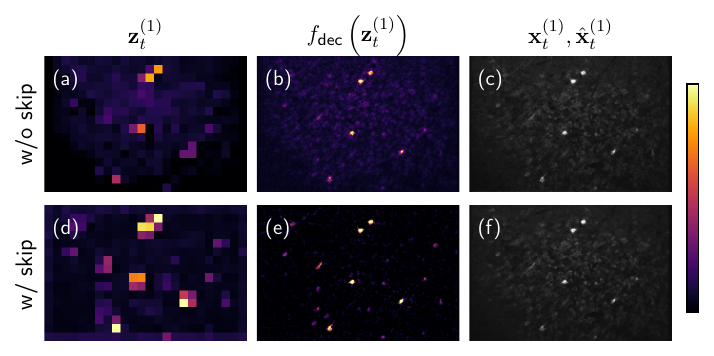}
    \caption{
      Effect of the skip connection on the latent representation and reconstruction from one run at $t=263$: 
      Top row \textbf{(a--c)}: model without skip connection. Bottom row \textbf{(d--f)}: model with skip connection.
      \textbf{(a,~d)}~Latent representation $\bm{z}_t^{(i)}$. 
      \textbf{(b,~e)}~Decoder output $f_{\text{dec}}(\bm{z}_t^{(i)})$, brightness enhanced for visual clarity.
      \textbf{(c)}~Ground truth input $\bm{x}_t^{(i)}$, contrast enhanced.
      \textbf{(f)}~Reconstruction $\bm{\hat{x}}_t^{(i)}$ with skip connection, contrast enhanced.
    }
    \label{fig:skip_connection}
\end{figure}

We introduce the skip connection to bypass static information directly to the decoder, preventing it from entering the latent space where it does not contribute temporal information conducive to \ac{VAR} modeling.
Figure \ref{fig:skip_connection} illustrates its effect on the latent representation.

The skip connection improves the signal-to-noise ratio of the latent space with respect to transient activations.
Comparing panels (a) and (d) in Figure \ref{fig:skip_connection}, the latent representation $\bm{z}_t^{(i)}$ without skip reflects both dynamic transients and static background structure.
While large activations remain visible, smaller transients are difficult to distinguish from the pervasive background.
With the skip connection, static content is routed through the aggregate frame to the decoder, causing it to vanish from the latent space.
This leaves $\bm{z}_t^{(i)}$ to represent primarily transient activations, where both strong and subtle temporal features now stand out clearly.

The effect is further illustrated by the decoder output $f_{\text{dec}}(\bm{z}_t^{(i)})$ in Panels (b) and (e). Without skip connection (Panel b), the latent space contains both dynamic and static content, and the decoder must reconstruct the full image from $\bm{z}_t^{(i)}$ alone.
With the skip connection (Panel e), the decoder renders localized 'blobs' corresponding to neural activity, while the static background is handled separately by the skip path. 

As a secondary benefit, reconstruction error $\mathcal{L}_{\text{rec}}$ decreases from $(9.20 \pm 2.52) \cdot 10^{-5}$ to $(8.08 \pm 2.80) \cdot 10^{-5}$ with the skip connection.

\subsection{Coefficients distinguish experimental conditions}
\label{sec:testing_results}


\begin{table}[ht]
  \centering
  \begin{tabular}{r@{ to }lr@{ to }l r@{.}l}
    \toprule
    \multicolumn{2}{c}{$G$}   & \multicolumn{2}{c}{$H$} & \multicolumn{2}{c}{p-value} \\
    \midrule
    $F_1$  & $F_{10}$  & $F_{11}$ & $F_{20}$  & 0 & 1781 \\
    $N_1$  & $N_{10}$  & $N_{11}$ & $N_{20}$  & 0 & 5136 \\
    $F_1$  & $F_{10}$  & $N_1$    & $N_{10}$  & {\bfseries 0} & {\bfseries 0002} \\
    $F_{11}$ & $F_{20}$ & $N_{11}$ & $N_{20}$  & {\bfseries 0} & {\bfseries 0021} \\
    $F_1$  & $F_{10}$  & $N_{11}$ & $N_{20}$  & {\bfseries 0} & {\bfseries 0001} \\
    $F_{11}$ & $F_{20}$ & $N_1$    & $N_{10}$  & {\bfseries 0} & {\bfseries 0002} \\
    \bottomrule
  \end{tabular}
  \caption{Group comparisons using VAR coefficients ($\lambda=0.005$), between runs F with familiar condition, between runs N with novel condition, and between runs F and N of different type. The similarity between sets of experimental runs based on their estimated dynamics is assessed by p-values. Significant p-values (after Bonferroni correction) are indicated by boldface.}
  \label{tab:testing}
\end{table}

To assess whether the \ac{VAR} coefficients $\bm{A}_1, ..., \bm{A}_p$ can uncover systematic differences in dynamic patterns between runs, we use the proposed testing procedure on the familiar (F) and novel (N) experimental conditions.
We performed pairwise similarity tests between these groups using the coefficients obtained from models trained with a Lasso penalty $\lambda = 0.005$.

The results are presented in Table \ref{tab:testing}.
When comparing subsets from the same experimental condition (e.g., early F runs vs. late F runs), the p-values were not statistically significant, after accounting for multiple comparisons.
Using a significance level of $\alpha = 0.05$ and applying Bonferroni correction for the $6$ tests performed, the observed within-condition p-values exceed the adjusted significance threshold.
This suggests that there are no strong differences between runs of the same condition, i.e. the estimated coefficients generalize across the runs.
In contrast, all comparisons between F and N conditions yielded significant p-values (all $p < 0.003$).
This indicates that the estimated \ac{VAR} coefficients capture the differences in neural dynamics between the conditions. Thus, the sparse \ac{VAR} coefficients derived by our end-to-end model seem to be meaningful descriptors of the underlying neural dynamics, capturing the differences between experimental conditions, while showing limited intra-condition variability.

\subsection{Interpreting the sparse coefficients}
\label{sec:coefficients}
\begin{figure}[!tb]
    \includegraphics[width=1\textwidth]{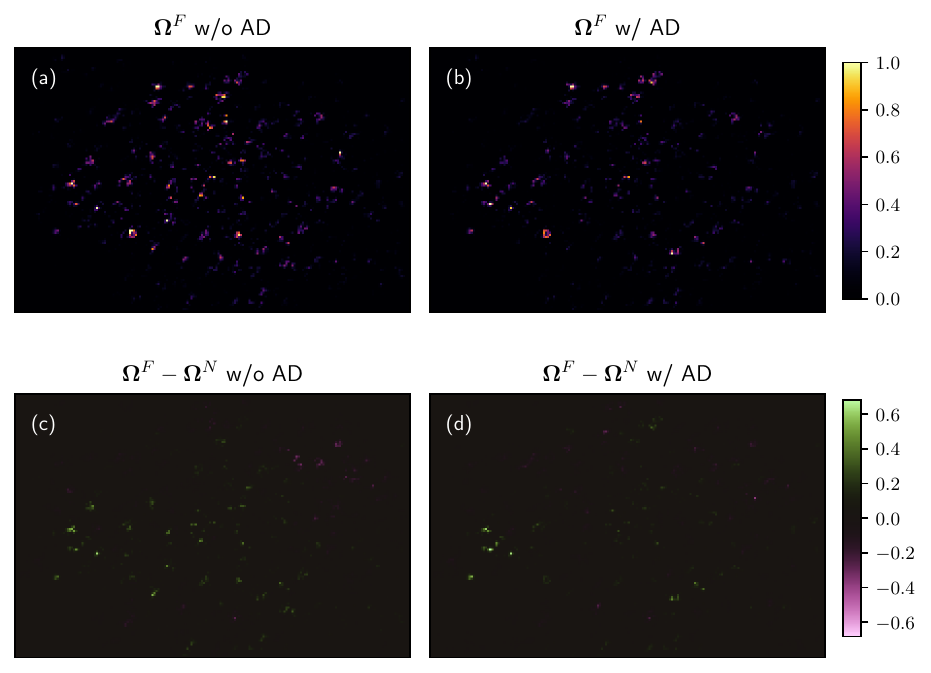}
    \caption{
       Assessing the effect end-to-end learning via automatic differentiation (AD) with visual contribution maps, $\bm{\Omega}$, which illustrate the spatial origins of estimated dynamics, averaged across all runs of an exemplary mouse: \textbf{(a)}~Familiar condition without end-to-end training. \textbf{(b)}~Familiar condition with end-to-end training. \textbf{(c)}~Difference map ($\bm{\Omega}^F - \bm{\Omega}^N$) without end-to-end training. \textbf{(d)}~Difference map ($\bm{\Omega}^F - \bm{\Omega}^N$) with end-to-end training. 
    }
    \label{fig:coeff_reconstructions}
\end{figure}

\begin{figure}[!tb]
    \includegraphics[width=1\textwidth]{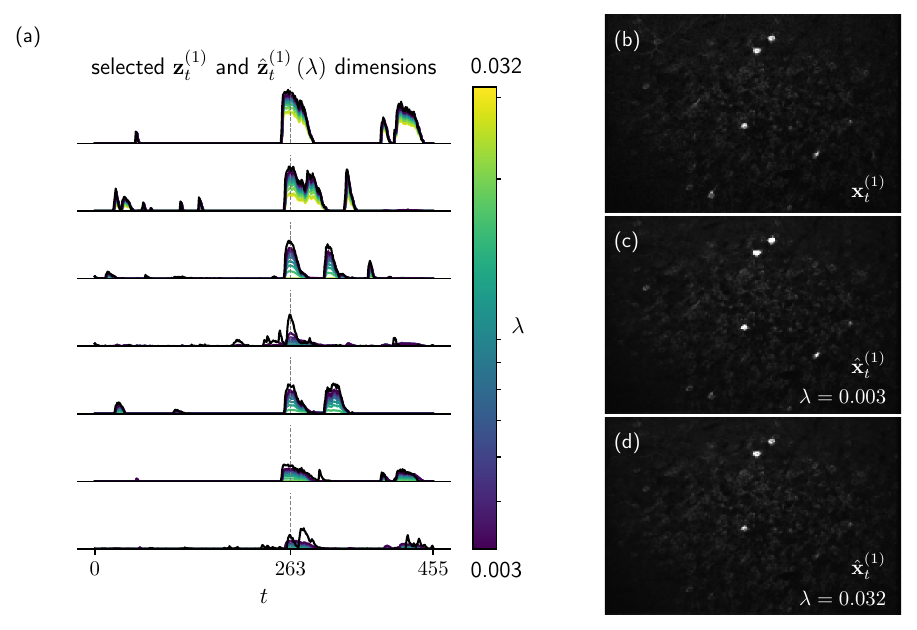}
    \caption{
Effect of $\ell_1$ regularization parameter $\lambda$ on time series forecast and reconstruction.
\textbf{(a)}~Selected dimensions of $\bm{z}_t$ in black (reference) and their forecasts $\hat{\bm{z}}_t$ for varying $\lambda$ (in colors) over time.
\textbf{(b--d)}~Reconstructions at $t=263$: \textbf{(b)}~Ground truth $\bm{x}_t$. \textbf{(c)}~Reconstruction $\hat{\bm{x}}_t$ for $\lambda = 0.003$. \textbf{(d)}~Reconstruction $\hat{\bm{x}}_t$ for $\lambda = 0.032$.
    }
    \label{fig:forecast_lambda}
\end{figure}

To inspect the estimated dynamics in more detail and identify the spatial origins of differences between experimental conditions, we compute the visual contribution maps $\bm{\Omega}^{(i)}$ for each time series $i$ and average them across all runs within the Familiar (F) and Novel (N) conditions, obtaining $\bm{\Omega}^{F}$ and $\bm{\Omega}^{N}$.
Figure \ref{fig:coeff_reconstructions} displays these maps, comparing training without end-to-end gradients (Panels a, c) to our proposed full end-to-end approach (Panels b, d).

The difference maps $\bm{\Omega}^F - \bm{\Omega}^N$ (Panels c, d) reveal clear differences between the experimental conditions.
The contribution map $\bm{\Omega}^F$ (Panels a, b) is predominantly positive, indicating higher outflow signal in the estimated coefficients in the familiar condition F across multiple neural assemblies.
This finding aligns with empirical evidence that familiar environments exhibit more stable and coordinated population activity \citep{sylte2025coordinated}: The higher outflow in F reflects stronger, more established influence relationships between neural assemblies. As seen from the difference maps, the novel condition N shows weaker structure, consistent with an exploratory phase where dynamics are less coordinated.

Comparing the two training strategies, our proposed end-to-end approach (Panels b, d) produces markedly sparser and more localized maps, revealing distinct, identifiable spatial structures.
In contrast, the embedded model without end-to-end gradients (Panels a, c) yields more diffuse and less interpretable patterns of influence. In particular,
end-to-end training (Panel d) produces a more distinct and localized pattern of the condition difference compared to training without backpropagation through \ac{LARS} (panel c).
These visual differences demonstrate how backpropagation through the \ac{LARS} solver shapes the learned representations to be more interpretable.

As a consistency check, we applied the proposed test at the level of influence vectors, comparing all F runs against all N runs.
Both training approaches yield significant differences: the embedded model gives $p = 0.026$ and the end-to-end model gives $p = 0.018$, confirming that the aggregated influence vectors $\bm{c}^{(i)}$ retain discriminative information from the full coefficient matrices.

We furthermore assessed the sensitivity of both the latent forecasts and the resulting reconstructions to the regularization parameter $\lambda$.
As shown in Figure \ref{fig:forecast_lambda}, lower regularization (e.g., $\lambda = 0.003$) produces forecasts that closely track the reference signal (Panel~a), and the corresponding reconstruction (Panel~c) closely matches the ground truth (Panel~b), capturing nearly all transient activations.
In contrast, higher regularization (e.g., $\lambda = 0.032$) leads to sparser forecasts that retain only the strongest activations (Panel~a), and the reconstruction (Panel~d) accordingly captures only the most prominent features while suppressing weaker transients.
This shows that $\lambda$ provides intuitive control over the trade-off between reconstruction fidelity and coefficient sparsity.
When tuning $\lambda$, visually inspecting the reconstruction or observing $\mathcal{L}_{\text{rec}}$ gives feedback on the desired level of regularization.
The approach works for a range of $\lambda$ values, and as intended, $\lambda$ effectively controls the sparsity of the learned coefficients. Yet, the identified structure and the reconstruction appear to be rather similar overall, at least in the example in Figure \ref{fig:forecast_lambda}, even when changing $\lambda$ by an order of a magnitude. Therefore, results might not be very sensitive to this tuning parameter, which could avoid the need for careful tuning.



\subsection{End-to-end training yields a more predictable latent space}
\label{sec:ablation}

To evaluate the effectiveness of our proposed end-to-end training strategy with differentiable \ac{LARS}, we conducted an ablation study comparing three different approaches for combining the autoencoder (AE) and the \ac{VAR} model, summarized in Table \ref{tab:ablation}.
The goal is to understand how the integration method impacts reconstruction error $\mathcal{L}_{\text{rec}}$ and the predictability of the latent space dynamics.
To assess the latter, we use the relative \ac{VAR} prediction error $\mathcal{R}_{\text{var}}$, which normalizes each time point's squared error by its squared signal magnitude:
$$\mathcal{R}_{\text{var}} = \frac{1}{N} \sum_{i=1}^{N} \frac{1}{T_i - p} \sum_{t=p+1}^{T_i} \frac{\|\bm{z}_t^{(i)} - \hat{\bm{z}}_t^{(i)}\|^2}{\|\bm{z}_t^{(i)}\|^2},$$
where $\hat{\bm{z}}_t^{(i)} = f_{\text{VAR}}(\bm{z}_{t-1}^{(i)},\ldots,\bm{z}_{t-p}^{(i)})$.

\begin{table}[t]
  \caption{Ablation study comparing different methods of combining the autoencoder and the VAR model (sequential training, embedded VAR without end-to-end gradients, and our proposed end-to-end approach) via reconstruction error ($\mathcal{L}_{\text{rec}}$) and prediction error $\mathcal{R}_{\text{var}}$ at three different levels of $\lambda$.}
  \label{tab:ablation}
  \centering
  \begin{tabular}{l l c c c}
    \toprule
    \multicolumn{2}{l}{} & Sequential & Embedded & End-to-End \\
    \midrule
    \multirow{2}{*}{$\lambda=0.005$} & \(\mathcal{L}_{\text{rec}}\) {\scriptsize[\(10^{-3}\)]} & 0.087 \(\pm\) 0.030 & 0.101 \(\pm\) 0.035 & 0.103 \(\pm\) 0.038 \\
 & \(\mathcal{R}_{\text{var}}\) & 0.472 \(\pm\) 0.040 & 0.421 \(\pm\) 0.054 & 0.398 \(\pm\) 0.068 \\
\hline
\multirow{2}{*}{$\lambda=0.01$} & \(\mathcal{L}_{\text{rec}}\) {\scriptsize[\(10^{-3}\)]} & 0.087 \(\pm\) 0.030 & 0.106 \(\pm\) 0.037 & 0.112 \(\pm\) 0.042 \\
 & \(\mathcal{R}_{\text{var}}\) & 0.682 \(\pm\) 0.063 & 0.611 \(\pm\) 0.074 & 0.530 \(\pm\) 0.081 \\
\hline
\multirow{2}{*}{$\lambda=0.02$} & \(\mathcal{L}_{\text{rec}}\) {\scriptsize[\(10^{-3}\)]} & 0.087 \(\pm\) 0.030 & 0.113 \(\pm\) 0.040 & 0.118 \(\pm\) 0.045 \\
 & \(\mathcal{R}_{\text{var}}\) & 0.853 \(\pm\) 0.061 & 0.791 \(\pm\) 0.086 & 0.699 \(\pm\) 0.072 \\
    \bottomrule
  \end{tabular}
\end{table}

A sequential baseline (left column of Table \ref{tab:ablation}), where the AE is trained first for reconstruction and the \ac{VAR} model is fitted afterwards, achieves the best reconstruction but the poorest latent predictability, as the latent space is not optimal for the \ac{VAR} task.

Embedding the \ac{VAR} model within the AE, using the \ac{VAR} forecast $\bm{\hat{z}}_t$ for reconstruction but without backpropagation through \ac{LARS}(middle column of Table \ref{tab:ablation}), significantly improves $\mathcal{R}_{\text{var}}$.
As a result, $\mathcal{L}_{\text{rec}}$ increases as the decoder reconstructs from the \ac{VAR} forecast $\bm{\hat{z}}_t$ rather than the original latent variable $\bm{z}_t$.
This configuration shows that even implicitly guiding the AE via the \ac{VAR} forecast helps shape a more predictable latent space.

Our full proposed method (right column of Table \ref{tab:ablation}), with end-to-end training via differentiable \ac{LARS}, produces a further reduction in $\mathcal{R}_{\text{var}}$, achieving the best latent predictability overall.
This key improvement, gained by allowing gradients to flow through the \ac{LARS} computation, comes at the cost of a further small increase in $\mathcal{L}_{\text{rec}}$. These results are consistent across a wide range of penalty parameter $\lambda$ values.

While embedding the \ac{VAR} model offers benefits over a sequential approach, the key step is enabling full end-to-end differentiation through the \ac{LARS} algorithm.
This allows the \ac{VAR} objective to directly shape the latent space, leading to a better capture of the dynamics by the sparse linear model.

\section{Discussion}
\label{sec:discussion}
In this work, we presented an end-to-end trainable framework that integrates a convolutional autoencoder with an $\ell_1$-penalized \ac{VAR} model, in which static components are addressed with a skip connection. Consequently, we combine the representational capabilities of artificial neural networks with the interpretability and theoretical grounding of sparse regression models.
The proposed framework produces interpretable, dimension-reduced representations of neural dynamics in an exemplary two-photon calcium imaging dataset, where a differentiable \ac{LARS} procedure enables optimal, sparse temporal latent spaces.

The three key innovations, the aggregate skip connection for separating static from dynamic content, the differentiable \ac{LARS} procedure, and the subsequent statistical testing approach, complement each other.
The skip connection frees the latent space from static clutter, while differentiable \ac{LARS} allows the encoder to directly optimize for predictable temporal dynamics.
An ablation study confirmed that both components contribute to performance: Embedding the \ac{VAR} model improves latent predictability over sequential training, and enabling gradient flow through \ac{LARS} produces further gains. 
The statistical testing approach then allows to assess the validity of patterns found in the data.

The sparsely estimated \ac{VAR} coefficients are meaningful descriptors of dynamics, e.g. capturing significant differences between experimental conditions in the exemplary application, while remaining robust to within-condition variability.
The proposed contribution map approach further indicate that end-to-end training produces sparser, more localized representations compared to non-differentiated approaches, thus improving interpretability.

Despite these promising results, our approach has several limitations.
The embedded \ac{LARS} fitting introduces computational complexity. We did not explore alternative $\ell_1$-solvers, which do not allow for gradients in a straightforward manner.
Also, the current architecture flattens the 2D latent maps before the \ac{VAR} model, potentially discarding spatial relationships that space-aware statistical models could exploit \citep{digiacinto2010vector}. While precise spatial relationships were less important in the exemplary application, as neurons may be connected via intermediate neurons that are not visible in the imaging plane, other applications might benefit from spatial information. For example, the VAR model could be adapted to only allow for connections up to a certain distance. Nevertheless, the present results already indicate that the differentiable \ac{LARS} method may prove valuable beyond this specific application, as it provides a general mechanism for integrating sparse regression into deep learning pipelines while avoiding the gradient conflicts and scaling issues of multi-task loss formulations \citep{sener2018multitask}.

Future work could refine the skip connection with adaptive mechanisms \citep{liu2019selfadaptive} or reduce latent dimensionality via differentiable top-k selection \citep{xie2020differentiable}.
More broadly, the successful differentiation of \ac{LARS} encourages exploring similar approaches for other numerical procedures, potentially benefiting domains such as climate modeling \citep{wesselkamp2024processinformed} or video analysis.
Surrogate gradients \citep{blondel2025elements} may further improve gradient quality through non-differentiable solver components.

More generally, our results show that differentiable programming techniques can bridge between artificial neural networks and sparse regression modeling for combining the benefits of both types of approaches.

\section*{Acknowledgments}
The work was funded by the Deutsche Forschungsgemeinschaft (DFG, German Research Foundation), FK, MH, JH, TB, AV, and HB via Project-ID 499552394 -- SFB 1597, MB and HB via Project-ID 514483642 -- TRR 384. AK was supported by the Hans A. Krebs Medical Scientist Program.

\bibliographystyle{unsrtnat}
\bibliography{processed}

@inproceedings{alaa2022icenode,
	title = {{{ICE-NODE}}: integration of clinical embeddings with neural ordinary differential equations},
	booktitle = {Proceedings of the 7th {{Machine Learning}} for {{Healthcare Conference}}},
	author = {Alaa, Asem and Mayer, Erik and Barahona, Mauricio},
	year = {2022},
	series = {Proceedings of {{Machine Learning Research}}},
	volume = {182},
	pages = {537--564},
	publisher = {PMLR}
}

@article{baydin2018automatic,
	title = {Automatic differentiation in machine learning: a survey},
	author = {Baydin, Atilim Gunes and Pearlmutter, Barak A. and Radul, Alexey Andreyevich and Siskind, Jeffrey Mark},
	year = {2018},
	journal = {Journal of Machine Learning Research},
	volume = {18},
	number = {153},
	pages = {1--43}
}

@article{beck2009fast,
	title = {A fast iterative shrinkage-thresholding algorithm for linear inverse problems},
	author = {Beck, Amir and Teboulle, Marc},
	year = {2009},
	journal = {SIAM Journal on Imaging Sciences},
	volume = {2},
	number = {1},
	pages = {183--202},
	doi = {10.1137/080716542}
}

@article{bengio1994learning,
	title = {Learning long-term dependencies with gradient descent is difficult},
	author = {Bengio, Yoshua and Simard, Patrice and Frasconi, Paolo},
	year = {1994},
	journal = {IEEE Transactions on Neural Networks},
	volume = {5},
	number = {2},
	pages = {157--166},
	doi = {10.1109/72.279181}
}

@inproceedings{bertrand2020implicit,
	title = {Implicit differentiation of {{Lasso-type}} models for hyperparameter optimization},
	booktitle = {Proceedings of the 37th {{International Conference}} on {{Machine Learning}}},
	author = {Bertrand, Quentin and Klopfenstein, Quentin and Blondel, Mathieu and Vaiter, Samuel and Gramfort, Alexandre and Salmon, Joseph},
	year = {2020},
	series = {Proceedings of {{Machine Learning Research}}},
	volume = {119},
	pages = {810--821},
	publisher = {PMLR}
}

@article{blondel2025elements,
	title = {The elements of differentiable programming},
	author = {Blondel, Mathieu and Roulet, Vincent},
	year = {2025},
	doi = {10.48550/arXiv.2403.14606},
	note = {arXiv preprint}
}

@inproceedings{bolte2021nonsmooth,
	title = {Nonsmooth implicit differentiation for machine-learning and optimization},
	booktitle = {Advances in {{Neural Information Processing Systems}}},
	author = {Bolte, J{\'e}r{\^o}me and Le, Tam and Pauwels, Edouard and {Silveti-Falls}, Tony},
	year = {2021},
	volume = {34},
	pages = {13537--13549},
	publisher = {Curran Associates, Inc.}
}

@book{boyd2004convex,
	title = {Convex {{Optimization}}},
	author = {Boyd, Stephen P. and Vandenberghe, Lieven},
	year = {2004},
	publisher = {Cambridge University Press}
}

@article{boyd2011distributed,
	title = {Distributed optimization and statistical learning via the alternating direction method of multipliers},
	author = {Boyd, Stephen and Parikh, Neal and Chu, Eric and Peleato, Borja and Eckstein, Jonathan},
	year = {2011},
	journal = {Foundations and Trends\textregistered{} in Machine Learning},
	volume = {3},
	number = {1},
	pages = {1--122},
	publisher = {Now Publishers, Inc.},
	doi = {10.1561/2200000016}
}

@inproceedings{casale2018gaussian,
	title = {Gaussian process prior variational autoencoders},
	booktitle = {Advances in {{Neural Information Processing Systems}}},
	author = {Casale, Francesco Paolo and Dalca, Adrian and Saglietti, Luca and Listgarten, Jennifer and Fusi, Nicolo},
	year = {2018},
	volume = {31},
	pages = {10390--10401},
	publisher = {Curran Associates, Inc.}
}

@book{casella2002statistical,
	title = {Statistical {{Inference}}},
	author = {Casella, George and Berger, Roger},
	year = {2002},
	edition = {2},
	publisher = {Duxbury},
	doi = {10.1201/9781003456285}
}

@article{cavalcante2017lasso,
	title = {Lasso vector autoregression structures for very short-term wind power forecasting},
	author = {Cavalcante, Laura and Bessa, Ricardo J. and Reis, Marisa and Browell, Jethro},
	year = {2017},
	journal = {Wind Energy},
	volume = {20},
	number = {4},
	pages = {657--675},
	doi = {10.1002/we.2029}
}

@inproceedings{chen2018gradnorm,
	title = {{{GradNorm}}: gradient normalization for adaptive loss balancing in deep multitask networks},
	booktitle = {Proceedings of the 35th {{International Conference}} on {{Machine Learning}}},
	author = {Chen, Zhao and Badrinarayanan, Vijay and Lee, Chen-Yu and Rabinovich, Andrew},
	year = {2018},
	series = {Proceedings of {{Machine Learning Research}}},
	volume = {80},
	pages = {794--803},
	publisher = {PMLR}
}

@inproceedings{chen2018neural,
	title = {Neural ordinary differential equations},
	booktitle = {Advances in {{Neural Information Processing Systems}}},
	author = {Chen, Ricky T. Q. and Rubanova, Yulia and Bettencourt, Jesse and Duvenaud, David K},
	year = {2018},
	volume = {31},
	pages = {6572--6583},
	publisher = {Curran Associates, Inc.}
}

@inproceedings{choromanska2015loss,
	title = {The loss surfaces of multilayer networks},
	booktitle = {Proceedings of the 18th {{International Conference}} on {{Artificial Intelligence}} and {{Statistics}}},
	author = {Choromanska, Anna and Henaff, {\relax Mi}kael and Mathieu, Michael and Ben Arous, Gerard and LeCun, Yann},
	year = {2015},
	series = {Proceedings of {{Machine Learning Research}}},
	volume = {38},
	pages = {192--204},
	publisher = {PMLR}
}

@inproceedings{couronne2021longitudinal,
	title = {Longitudinal self-supervision to~disentangle inter-patient variability from~disease progression},
	booktitle = {Medical {{Image Computing}} and {{Computer Assisted Intervention}} -- {{MICCAI}} 2021},
	author = {Couronn{\'e}, Rapha{\"e}l and Vernhet, Paul and Durrleman, Stanley},
	year = {2021},
	series = {Lecture {{Notes}} in {{Computer Science}}},
	volume = {12902},
	pages = {231--241},
	publisher = {Springer},
	doi = {10.1007/978-3-030-87196-3_22}
}

@article{digiacinto2010vector,
	title = {On vector autoregressive modeling in space and time},
	author = {Di Giacinto, Valter},
	year = {2010},
	journal = {Journal of Geographical Systems},
	volume = {12},
	number = {2},
	pages = {125--154},
	doi = {10.1007/s10109-010-0116-6}
}

@article{efron2004least,
	title = {Least angle regression},
	author = {Efron, Bradley and Hastie, Trevor and Johnstone, Iain and Tibshirani, Robert},
	year = {2004},
	journal = {The Annals of Statistics},
	volume = {32},
	number = {2},
	doi = {10.1214/009053604000000067}
}

@inproceedings{fortuin2020gpvae,
	title = {{{GP-VAE}}: deep probabilistic time series imputation},
	booktitle = {Proceedings of the 23rd {{International Conference}} on {{Artificial Intelligence}} and {{Statistics}}},
	author = {Fortuin, Vincent and Baranchuk, Dmitry and Raetsch, Gunnar and Mandt, Stephan},
	year = {2020},
	series = {Proceedings of {{Machine Learning Research}}},
	volume = {108},
	pages = {1651--1661},
	publisher = {PMLR}
}

@article{friedman2010regularization,
	title = {Regularization paths for generalized linear models via coordinate descent},
	author = {Friedman, Jerome H. and Hastie, Trevor and Tibshirani, Rob},
	year = {2010},
	journal = {Journal of Statistical Software},
	volume = {33},
	pages = {1--22},
	doi = {10.18637/jss.v033.i01}
}

@article{grienberger2022twophoton,
	title = {Two-photon calcium imaging of neuronal activity},
	author = {Grienberger, Christine and Giovannucci, Andrea and Zeiger, William and {Portera-Cailliau}, Carlos},
	year = {2022},
	journal = {Nature Reviews Methods Primers},
	volume = {2},
	number = {1},
	pages = {67},
	doi = {10.1038/s43586-022-00147-1}
}

@article{haselimashhadi2019differentiable,
	title = {A differentiable alternative to the {{Lasso}} penalty},
	author = {Haselimashhadi, Hamed and Vinciotti, Veronica},
	year = {2019},
	doi = {10.1080/03610926.2018.1515362},
	note = {arXiv preprint}
}

@inproceedings{kim2018interpretability,
	title = {Interpretability beyond feature attribution: quantitative testing with concept activation vectors ({{TCAV}})},
	booktitle = {Proceedings of the 35th {{International Conference}} on {{Machine Learning}}},
	author = {Kim, Been and Wattenberg, Martin and Gilmer, Justin and Cai, Carrie and Wexler, James and Viegas, Fernanda and Sayres, Rory},
	year = {2018},
	series = {Proceedings of {{Machine Learning Research}}},
	volume = {80},
	pages = {2668--2677},
	publisher = {PMLR}
}

@inproceedings{kingma2015adam,
	title = {Adam: a method for stochastic optimization},
	booktitle = {Proceedings of the 3rd {{International Conference}} on {{Learning Representations}} ({{ICLR}})},
	author = {Kingma, Diederik P. and Ba, Jimmy},
	year = {2015},
	bibsource = {dblp computer science bibliography, https://dblp.org}
}

@article{kober2022individualizing,
	title = {Individualizing deep dynamic models for psychological resilience data},
	author = {K{\"o}ber, G{\"o}ran and Pooseh, Shakoor and Engen, Haakon and Chmitorz, Andrea and Kampa, Miriam and Schick, Anita and Sebastian, Alexandra and T{\"u}scher, Oliver and Wessa, Mich{\`e}le and Yuen, Kenneth S. L. and Walter, Henrik and Kalisch, Raffael and Timmer, Jens and Binder, Harald},
	year = {2022},
	journal = {Scientific Reports},
	volume = {12},
	number = {1},
	pages = {8061},
	publisher = {Nature Publishing Group},
	doi = {10.1038/s41598-022-11650-6}
}

@inproceedings{krishnan2017structured,
	title = {Structured inference networks for nonlinear state space models},
	booktitle = {Proceedings of the {{AAAI Conference}} on {{Artificial Intelligence}}},
	author = {Krishnan, Rahul and Shalit, Uri and Sontag, David},
	year = {2017},
	volume = {31},
	doi = {10.1609/aaai.v31i1.10779}
}

@inproceedings{li2018visualizing,
	title = {Visualizing the loss landscape of neural nets},
	booktitle = {Advances in {{Neural Information Processing Systems}}},
	author = {Li, Hao and Xu, Zheng and Taylor, Gavin and Studer, Christoph and Goldstein, Tom},
	year = {2018},
	volume = {31},
	publisher = {Curran Associates, Inc.}
}

@inproceedings{liu2019selfadaptive,
	title = {Self-adaptive scaling for learnable residual structure},
	booktitle = {Proceedings of the 23rd {{Conference}} on {{Computational Natural Language Learning}} ({{CoNLL}})},
	author = {Liu, Fenglin and Gao, Meng and Liu, Yuanxin and Lei, Kai},
	year = {2019},
	pages = {862--870},
	publisher = {Association for Computational Linguistics},
	doi = {10.18653/v1/K19-1080}
}

@book{lutkepohl2005new,
	title = {New {{Introduction}} to {{Multiple Time Series Analysis}}},
	author = {L{\"u}tkepohl, Helmut},
	year = {2005},
	publisher = {Springer}
}

@article{pachitariu2017suite2p,
	title = {Suite2p: beyond 10,000 neurons with standard two-photon microscopy},
	author = {Pachitariu, Marius and Stringer, Carsen and Dipoppa, Mario and Schr{\"o}der, Sylvia and Rossi, L. Federico and Dalgleish, Henry and Carandini, Matteo and Harris, Kenneth D.},
	year = {2017},
	doi = {10.1101/061507},
	note = {bioRxiv preprint}
}

@article{pnevmatikakis2019analysis,
	title = {Analysis pipelines for calcium imaging data},
	author = {Pnevmatikakis, Eftychios A},
	year = {2019},
	journal = {Current Opinion in Neurobiology},
	volume = {55},
	pages = {15--21},
	doi = {10.1016/j.conb.2018.11.004}
}

@inproceedings{ramchandran2021longitudinal,
	title = {Longitudinal variational autoencoder},
	booktitle = {Proceedings of the 24th {{International Conference}} on {{Artificial Intelligence}} and {{Statistics}}},
	author = {Ramchandran, Siddharth and Tikhonov, Gleb and Kujanp{\"a}{\"a}, Kalle and Koskinen, Miika and L{\"a}hdesm{\"a}ki, Harri},
	year = {2021},
	series = {Proceedings of {{Machine Learning Research}}},
	volume = {130},
	pages = {3898--3906},
	publisher = {PMLR}
}

@inproceedings{rubanova2019latent,
	title = {Latent ordinary differential equations for irregularly-sampled time series},
	booktitle = {Advances in {{Neural Information Processing Systems}}},
	author = {Rubanova, Yulia and Chen, Ricky T. Q. and Duvenaud, David K},
	year = {2019},
	volume = {32},
	pages = {5321--5331},
	publisher = {Curran Associates, Inc.}
}

@inproceedings{sauty2022progression,
	title = {Progression models for~imaging data with~longitudinal variational auto encoders},
	booktitle = {Medical {{Image Computing}} and {{Computer Assisted Intervention}} -- {{MICCAI}} 2022},
	author = {Sauty, Beno{\^i}t and Durrleman, Stanley},
	year = {2022},
	series = {Lecture {{Notes}} in {{Computer Science}}},
	volume = {13431},
	pages = {3--13},
	publisher = {Springer},
	doi = {10.1007/978-3-031-16431-6_1}
}

@inproceedings{sener2018multitask,
	title = {Multi-task learning as multi-objective optimization},
	booktitle = {Advances in {{Neural Information Processing Systems}}},
	author = {Sener, Ozan and Koltun, Vladlen},
	year = {2018},
	volume = {31},
	pages = {525--536},
	publisher = {Curran Associates, Inc.}
}

@article{shlezinger2023modelbased,
	title = {Model-based deep learning},
	author = {Shlezinger, Nir and Whang, Jay and Eldar, Yonina C. and Dimakis, Alexandros G.},
	year = {2023},
	journal = {Proceedings of the IEEE},
	volume = {111},
	number = {5},
	pages = {465--499},
	doi = {10.1109/JPROC.2023.3247480}
}

@misc{sylte2025coordinated,
	title = {Coordinated representational drift supports stable place coding in hippocampal {{CA1}}},
	author = {Sylte, Ole Christian and Kilias, Antje and Bartos, Marlene and Sauer, Jonas-Frederic},
	year = {2025},
	publisher = {Neuroscience},
	doi = {10.1101/2025.02.04.636428}
}

@article{tibshirani1996regression,
	title = {Regression shrinkage and selection via the {{Lasso}}},
	author = {Tibshirani, Robert},
	year = {1996},
	journal = {Journal of the Royal Statistical Society. Series B (Methodological)},
	volume = {58},
	number = {1},
	pages = {267--288},
	publisher = {[Royal Statistical Society, Oxford University Press]},
	doi = {10.1111/j.2517-6161.1996.tb02080.x}
}

@inproceedings{watter2015embed,
	title = {Embed to control: a locally linear latent dynamics model for control from raw images},
	booktitle = {Advances in {{Neural Information Processing Systems}}},
	author = {Watter, Manuel and Springenberg, Jost and Boedecker, Joschka and Riedmiller, Martin},
	year = {2015},
	volume = {28},
	publisher = {Curran Associates, Inc.}
}

@article{wesselkamp2024processinformed,
	title = {Process-informed neural networks: a hybrid modelling approach to improve predictive performance and inference of neural networks in ecology and beyond},
	author = {Wesselkamp, Marieke and Moser, Niklas and Kalweit, Maria and Boedecker, Joschka and Dormann, Carsten F.},
	year = {2024},
	journal = {Ecology Letters},
	volume = {27},
	number = {11},
	pages = {e70012},
	doi = {10.1111/ele.70012}
}

@article{wilcoxon1945individual,
	title = {Individual comparisons by ranking methods},
	author = {Wilcoxon, Frank},
	year = {1945},
	journal = {Biometrics Bulletin},
	volume = {1},
	number = {6},
	pages = {80},
	doi = {10.2307/3001968}
}

@article{wu2008coordinate,
	title = {Coordinate descent algorithms for {{Lasso}} penalized regression},
	author = {Wu, Tong Tong and Lange, Kenneth},
	year = {2008},
	journal = {The Annals of Applied Statistics},
	volume = {2},
	number = {1},
	doi = {10.1214/07-AOAS147}
}

@inproceedings{xie2020differentiable,
	title = {Differentiable top-k with optimal transport},
	booktitle = {Advances in {{Neural Information Processing Systems}}},
	author = {Xie, Yujia and Dai, Hanjun and Chen, Minshuo and Dai, Bo and Zhao, Tuo and Zha, Hongyuan and Wei, Wei and Pfister, Tomas},
	year = {2020},
	volume = {33},
	pages = {20520--20531},
	publisher = {Curran Associates, Inc.}
}

@inproceedings{yang2016deep,
	title = {Deep {{ADMM-Net}} for compressive sensing {{MRI}}},
	booktitle = {Advances in {{Neural Information Processing Systems}}},
	author = {Yang, Yan and Sun, Jian and Li, Huibin and Xu, Zongben},
	year = {2016},
	volume = {29},
	pages = {10--18},
	publisher = {Curran Associates, Inc.}
}

\end{document}